\documentclass{article}
\usepackage{spconf,amsmath}
\usepackage{amsmath,amsfonts}
\usepackage{algorithmic}
\usepackage{array}
\usepackage{subcaption}
\usepackage{textcomp}
\usepackage{stfloats}
\usepackage{url}
\usepackage{verbatim}
\usepackage{graphicx}

\usepackage{hyperref}
\hypersetup{hidelinks}

\usepackage{comment}
\usepackage{booktabs}
\usepackage{multirow}
\usepackage[flushleft]{threeparttable}
\usepackage{makecell}
\title{ViTransPAD: Video Transformer using convolution and self-attention\\for Face Presentation Attack Detection}
\name{\makecell{Zuheng~Ming$^{1}$,  Zitong~Yu$^{2}$,  Musab Al-Ghadi$^{1}$, \\ Muriel Visani$^{1,3}$ , Muhammad MuzzamilLuqman$^{1}$ , Jean-Christophe Burie$^{1}$} 
   }
\address{
\normalsize{$^{1}$~L3i, University of La Rochelle, La Rochelle, France}
\quad  \normalsize{$^{2}$~CMVS, University of Oulu, Finland}\\
\normalsize{$^{3}$~School of Information \& Communication Technology, Hanoi University of Science and Technology, Vietnam}\\
\small{\{zuheng.ming,musab.alghadi,muriel.visani, muhammad\_muzzamil.luqman, jcburie\}@univ-lr.fr \quad zitong.yu@oulu.fi }}

\begin{document}
%\ninept
%
\maketitle

\begin{abstract}
Face Presentation Attack Detection (PAD) is an important measure to prevent spoof attacks for face biometric systems.
Many works based on Convolution Neural Networks (CNNs) for face PAD formulate the problem as an image-level binary classification task without considering the context. Alternatively,  Vision Transformers (ViT) using self-attention to attend the context of an image become the mainstreams in face PAD. Inspired by ViT, we propose a Video-based Transformer for face PAD (ViTransPAD) with short/long-range spatio-temporal attention which can not only focus on local details with short attention within a frame but also capture long-range dependencies over frames. 
Instead of using coarse image patches with single-scale as in ViT, we propose the Multi-scale Multi-Head Self-Attention (MsMHSA) architecture to accommodate  multi-scale patch partitions of Q, K, V feature maps to the heads of transformer in a coarse-to-fine manner, which enables to learn a fine-grained representation to perform pixel-level discrimination for face PAD. 
Due to lack inductive biases of convolutions in pure transformers, we also introduce convolutions to the proposed ViTransPAD to integrate the desirable properties of CNNs by using convolution patch embedding and convolution projection. The extensive experiments show the effectiveness of our proposed ViTransPAD with a preferable accuracy-computation balance, which can serve as a new backbone for face PAD.
\end{abstract}
\begin{keywords}
Video-based transformer, multi-scale multi-head self-attention, face presentation attack detection
\end{keywords}
%\begin{keywords}
%Video-based transformer, multi-scale %multi-head self-attention, convolution %neural networks, face PAD
%\end{keywords}
%
\section{Introduction}
\label{sec:intro}

\begin{figure}
  \includegraphics[scale=0.64]{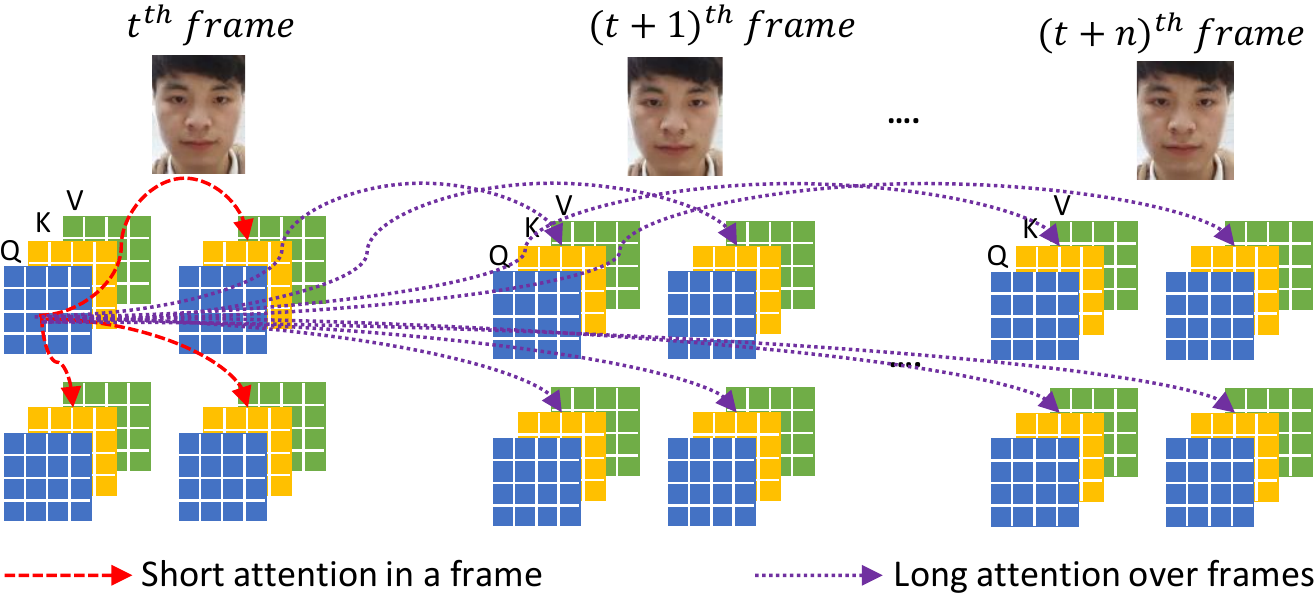}
  \caption{The short/long-range spatio-temporal attention of the proposed ViTransPAD, which can not only focus on local spatial details with short attention within a frame but also capture long-range temporal dependencies over frames.}
  \label{fig:ViTransAttent}
\end{figure}

%The breakthrough of biometrics authentication via deep learning offers a practical solution to authentication-required applications.
%The vulnerability of face biometric systems has been widely recognized and brought increasing attention to the vision community due to various physical and digital attacks~\cite{liu2020disentangling}. 
Face Presentation Attack Detection (PAD)~\cite{yu2021deep} is an important measure to prevent spoof attacks for biometric user authentication when using face biometric systems. Many works based on Convolution Neural Networks (CNNs) for face PAD formulate the problem as an image-level binary classification task to distinguish the bona fide from Presentation Attacks (PAs)~\cite{liu2020disentangling,yu2021}.

%on the input images or the estimated pseudo-depth maps as 
The image-based methods are simple and high-efficient. Nevertheless, these methods neglect the context information being useful to improve the generalization performance of face PAD models~\cite{wang2020deep}. 
%Some approach based on 3D convolution has been also proposed to learn the spatio-temporal context for distinguishing the live from spoof images~\cite{gan20173d}. 
Due to the limited receptive field, 3D convolution-based face PAD~\cite{gan20173d} also suffers from difficulty in learning long-range dependency.

Alternatively,  
%the recent Vision Transformers (ViT)~\cite{dosovitskiy2020image} using self-attention transformer allows to attend the global context of a given patch in an image.  
Vision Transformers (ViT)~\cite{dosovitskiy2020image} using self-attention to attend the global context of an image is becoming a new mainstream in face PAD~\cite{george2021effectiveness,yu2021transrppg}. 
%due to their strong semantic representation capacities to distinguish the bonafide from Presentation Attacks (PAs)~\cite{yu2022flexible}. 
However, ViT is also incapable to model the long-range dependencies over all frames in a video~\cite{arnab2021vivit}.   

Inspired by ViT, we propose a Video-based Transformer for face PAD (ViTransPAD) with short/long-range spatio-temporal attention, which can not only focus on local spatial details with short attention within a frame but also model long-range temporal dependencies over frames (see~\figureautorefname~\ref{fig:ViTransAttent}). The visualisation of attention maps (in Section~\ref{sec:visualisation}) shows that the proposed ViTransPAD based on short/long-range dependencies can gain a consistent attention being less affected by the noise. 
%such as illumination variation and head pose for face PAD. 
Instead of factorizing the spatio-temporal attention~\cite{arnab2021vivit}, we jointly learn spatio-temporal dependencies in our ViTransPAD.

%JCB
%Due to use coarse image patches with single-scale as input, the vanilla ViT is difficult to be directly adapted to the pixel-level dense prediction tasks such as object detection and segmentation~\cite{wang2021pyramid} as well as face PAD.
Due to the use of coarse image patches with a single input scale, it is difficult to directly adapt the vanilla ViT to the pixel-level dense prediction tasks such as object detection and segmentation~\cite{wang2021pyramid} as well as face PAD. 
To address this problem,  we propose the Multi-scale Multi-Head Self-Attention (MsMHSA) architecture 
% JCB
%to accommodate multi-scale patch partitioned
to support partitioned multi-scale patches
 from Q, K, V feature maps to different heads of transformer in a coarse-to-fine manner, which allows to learn a fine-grained representation to perform pixel-level discrimination required by face PAD. Rather than hierarchically stacking  multiple transformers as in~\cite{wang2021pyramid}, we simply implement MsMHSA in a single transformer to attain a preferable computation-accuracy balance. 
%Rather than stacking transformers as in~\cite{wang2021pyramid}, the proposed MsMHSA is simply conducted in a single transformer with less computation complexity. 

Nevertheless, the pure transformers such as ViT lack some of the inductive biases of convolutions requiring more data to train the models~\cite{wu2021cvt} which is not suitable to face PAD training on relative small datasets~\cite{Boulkenafet2017OULU,Zhang2012A,ReplayAttack,wen2015face}. To address this problem, we introduce convolutions to our ViTransPAD to tokenise video and employ convolutional projection rather than linear projection to encode Q, K, V feature maps for self-attention. The integrated CNNs force to capture the local spatial structure which allows to drop positional encoding being crucial for pure transformers.
%to encode the spatial information.

To summarize, the main contributions of this work are: 1) The design of a Video-based Transformer for face PAD (ViTransPAD) with short/long-range spatio-temporal attention. 2) A Multi-scale Multi-Head Self-Attention (MsMHSA) implemented on single transformer allowing to perform pixel-level fine-grained classification with good computation-accuracy balance for face PAD. 3) The introduction of convolutions to proposed ViTransPAD to integrate desirable proprieties. 4) To the best of our knowledge, this is the first approach using video-based transformer for face PAD. The superior performance demonstrates that the proposed ViTransPAD can serve as an effective new backbone for face PAD.

\section{Related works}
\label{sec:relatedwork}

\noindent\textbf{Face presentation attack detection.}\quad Traditional face PAD methods usually extract hand-crafted descriptors from the facial images to capture the different clues  such as liveness clues (including remote photoplethysmography (rPPG)), texture clues and 3D geometric clues to defend against photo print/video replay/3D masks attacks~\cite{ming2020survey}. 
Then, deep learning based methods using CNNs learn the representations of different clues from the images to distinguish the bona fide from PAs~\cite{yu2021, Liu2018Learning}. Recently,  ViT using self-attention is becoming a new mainstream in face PAD \cite{george2021effectiveness} due to their strong semantic representation capacities to detect PAs. 
%Nevertheless, Vit only attends the local structure in an image but not context over frames in a video. 
Some methods based on 3D CNNs~\cite{gan20173d} or 2D CNNs~\cite{wang2020deep,yu2020face,yu2020fas} with auxiliary plugging components try to model long-range dependencies for face PAD. However, the limited receptive field of 3D CNNs hinders its capacity to learn long-range context. 
\noindent\textbf{Vision transformers.}\quad 
%Transformer has gained a great success in the field of natural language processing (NLP)~\cite{vaswani2017attention}. 
The first work to introduce Transformer~\cite{vaswani2017attention} to vision domain is Vision Transformers(ViT)~\cite{dosovitskiy2020image}. 
%which adapts the encoder of original Transformer to image input. 
Thanks to self-attention, ViT and the variants ~\cite{liu2021swin,touvron2021training} show their superiority in image classification and in downstream tasks~\cite{ zhu2020deformable}. However, these image-based vision transformers only consider the spatial attention in a frame without integrating temporal attention over frames. ViViT~\cite{arnab2021vivit} model the long-range dependencies over frames with pure transformers. In order to introduce inductive bias of convolutions in pure transformers, 
%JCB : 
%convolution patch embedding  is proposed to use in ~\cite{wu2021cvt}.
%
~\cite{wu2021cvt} propose to use embedded convolution patches.
Multi-scale patches are applied in hierarchical stacking transformers~\cite{wang2021pyramid}  to adapt vanilla ViT to pixel-level dense prediction tasks. In this work, we design a simple MsMHSA architecture within a single video-based transformer allowing pixel-level fine-grained discrimination to satisfy the requirement of face PAD.

\section{Methodology}
\label{sec:Method}

\subsection{Overall Architecture}
\label{sec:Architecture}

\begin{figure*}
  \includegraphics[scale=0.52]{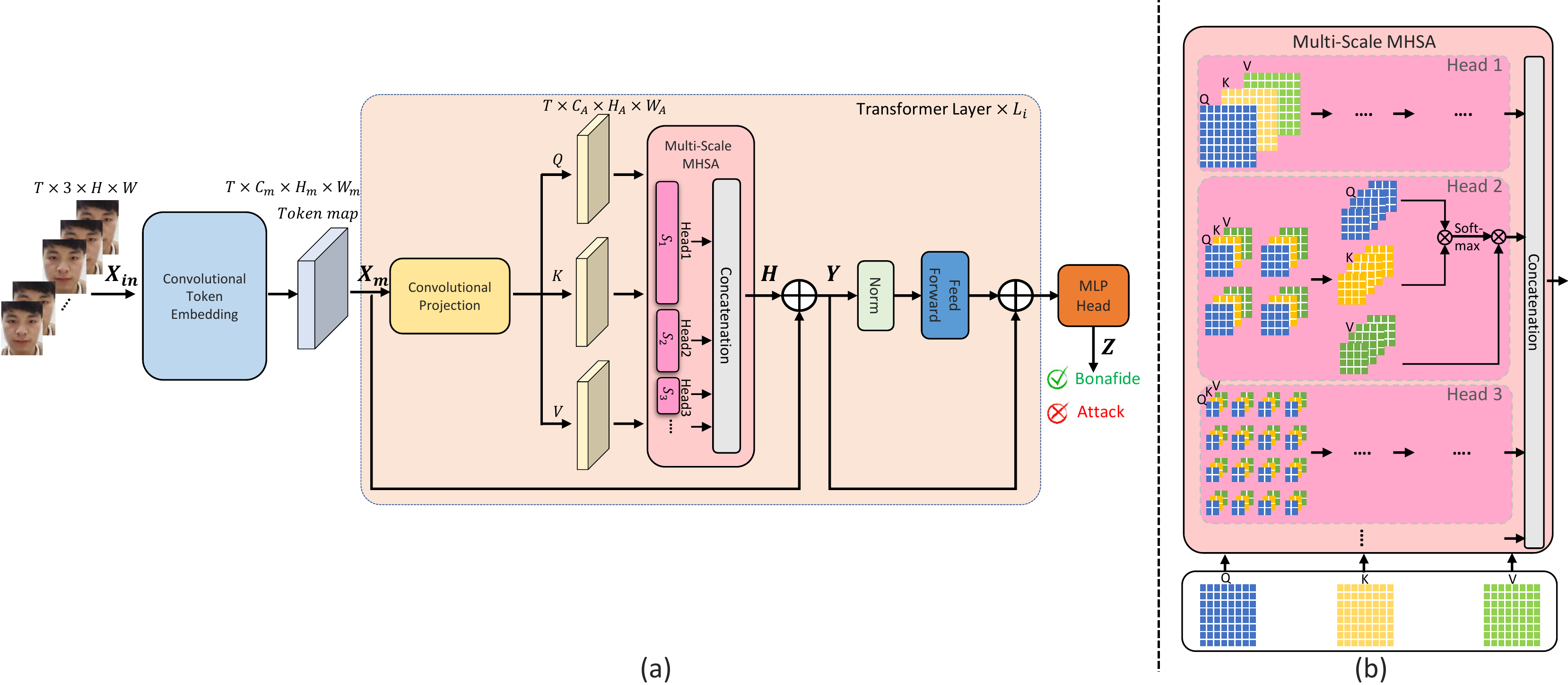}
  \caption{(a) Overall architecture of the Video-based Transformer for face PAD (ViTransPAD). (b) The proposed Multi-scale Multi-Head Self-Attention (MsMHSA) module in our ViTransPAD.}
  \label{fig:ViTransPAD}
\end{figure*}

An overview of the proposed ViTransPAD is depicted in~\figureautorefname~\ref{fig:ViTransPAD} (a). Unlike hierarchical stacking transformers using multi-scale patches in~\cite{wang2021pyramid}, we only use a single transformer to apply the multi-scale self-attention with different heads in each layer of transformer.

%\vspace{0.3em}
\noindent\textbf{\textit{Convolutional token embedding (CTE)}}.\quad Instead of partitioning each frame into patches and then tokenize patches with linear projection layer, we introduce a convolutional layer to tokenize each frame without partition. Given an input video $\mathbf{X_{in}}$ of size $T \times 3 \times H \times W$, the obtained token map is $\mathbf{X_m} \in\mathbb{R}^{T \times C_m \times H_m \times W_m}$, where $T$ is the number of frames.
%and $C_m, H_m, W_m$ are the channel, height and width of token map. Since the convolutional operation forces to encode the spatial structure, the positional encoding used in pure transformer as ViT is not necessary here.  

%\vspace{0.3em}
\noindent\textbf{\textit{Convolutional projection (CP)}}.\quad As well as CTE, we use  convolutional projection rather than linear projection to encode $\mathbf{Q}/\mathbf{K}/\mathbf{V}\in\mathbb{R}^{T \times C_A \times H_A \times W_A}$ feature maps. 
%\vspace{0.3em}
%using convolutional projection rather than linear projection used in pure transformers. 
%In practice, three pointwise convolution~\cite{howard2017mobilenets} are applied to embed the feature maps.  The pointwise convolution can much reduce computation cost of transformers only slightly degrading performance~\cite{li2022uniformer}. 

Then the obtained $\mathbf{Q}$, $\mathbf{K}$ and $\mathbf{V}$ are fed into the proposed Multi-scale Multi-Heads Self-Attention (MsMHSA) module to learn the short/long-range dependencies over frames in a video. The details are described in Section~\ref{sec:MsMHSA}. Finally, we add Feed-Forward Network (FFN) with Norm layers at the end of transformer. In this work, linear projection layers in FFN are also replaced by convolution layers. As in ViT, an MLP Head is connected to the transformer to generate the classification embeddings $\mathbf{Z}$ for face PAD. Given an input video $\mathbf{X_{in}}$, 
the proposed ViTransPAD can be described as:
\begin{equation}
\label{CTE} 
\mathbf{X_m} = \mathbf{CTE(X_{in})}
\end{equation}

\begin{equation}
\label{CP} 
\mathbf{Q, K, V} = \mathbf{CP(X_{m})}
\end{equation}

\begin{equation}
\label{MsMHSA} 
\mathbf{H} = \mathbf{MsMHSA(Q, K, V)}
\end{equation}

\begin{equation}
\label{MsMHSA} 
\mathbf{Y} = \mathbf{H+X_m}
\end{equation}

\begin{equation}
\label{FFN} 
\mathbf{Z} = \mathbf{MLP(FFN(Norm(Y))+Y)}
\end{equation}

\subsection{Multi-scale Multi-Head Self-Attention (MsMHSA)}
\label{sec:MsMHSA}
The goal of MsMHSA, as shown in~\figureautorefname~\ref{fig:ViTransPAD}(b), is to introduce a pyramid structure into the self-attention module to generate multi-scale feature maps which can be used for pixel-level fine-grained image discrimination required by face PAD. The proposed MsMHSA is applied on different heads of each layer of our transformer. All the heads share the similar protocol to calculate the self-attention. 
%The only difference is that the size of input patches partitioned from the $\mathbf{Q/K/V}$ feature maps varies in a coarse-to-fine manner. 
In particular, the  feature maps $\mathbf{Q/K/V}$  are equally divided to each head along the dimension before inputting them to MsMHSA. 
%(the $\mathbf{Q/K/V}$ for the last head may have less dimension if the dimension cannot be evenly divided).

Given a transformer with three heads fed by an input feature maps $\mathbf{Q/K/V}$ of size $T \times C_A \times H_A \times W_A$, the feature maps for each head are $\mathbf{Q_i/K_i/V_i}\in \mathbb{R}^{T \times \frac{C_A}{3}\times H_A \times W_A} $. For the first head $Head_1$, we take a full-size patch $\mathbf{q_1/k_1/v_1 }\in\mathbb{R}^{T \times \frac{C_A}{3} \times H_A \times W_A}$ to calculate a global self-attention feature map for the first head $Head_1$. We can obtain the self-attention feature map $\mathbf{h_1}\in\mathbb{R}^{T \times \frac{C_A}{3} \times H_A \times W_A}$ of $Head_1$. Then for $Head_2$, we divide $\mathbf{Q_2/K_2/V_2}$ into $2^2$ patches, each patch $\mathbf{q_2/k_2/v_2}$ of size ${T \times \frac{C_A}{3} \times \frac{H_A}{2} \times \frac{W_A}{2}}$. We  obtain the self-attention feature map $\mathbf{h_2}\in\mathbb{R}^{T \times \frac{C_A}{3} \times \frac{H_A}{2} \times \frac{W_A}{2}}$ of $Head_2$. We continue to divide $\mathbf{Q_3/K_3/V_3}$ into $4^2$ patches to calculate the self-attention feature map $\mathbf{h_3}\in\mathbb{R}^{T \times \frac{C_A}{3} \times \frac{H_A}{4} \times \frac{W_A}{4}}$ of $Head_3$. Finally, we concatenate the obtained self-attention feature maps $\{\mathbf{h_1},\mathbf{h_2},\mathbf{h_3}\}$ to generate the final multi-scale attention feature map $\mathbf{H}\in\mathbb{R}^{T \times C_A \times H_A \times W_A}$ in layer $L_i$ (We need to reshape $\mathbf{h_2},\mathbf{h_3}$ to be consistent with $\mathbf{h_1}$):
\begin{equation}
\label{MsMHSA} 
\mathbf{H} = \mathbf{Concat(\mathbf{h_1}, \mathbf{Reshape(h_2)}, \mathbf{Reshape(h_3)})}.
\end{equation}
%\vspace{-0.8em}
The self-attention feature map $\mathbf{h_i}$ is given by:
%%\vspace{-0.5em}
\begin{equation}
\label{MsMHSA} 
\mathbf{h_i} = \Sigma_m^N\Sigma_n^N{\mathbf{Softmax}(\frac{\mathbf{q_{i,m}k_{i,n}^T}}{\sqrt{d_{head_{i,n}}}})\mathbf{v_{i,n}}},
\end{equation}

where $i$ is corresponding to the $i$th head $Head_i$, $\mathbf{q_{i,m}}$ is the $m$th patch partitioned from feature map $\mathbf{Q_i}$ for the  $i$th head $Head_i$, $\mathbf{k_{i,n}/v_{i,n}}$ are the $n$th patches partitioned from feature maps $\mathbf{K_i/V_i}$ for the $i$th head $Head_i$, then, $\mathbf{q_{i,m}/k_{i,n}/v_{i,n} }\in\mathbb{R}^{\frac{C_A}{3} \times \frac{W_A}{l} \times \frac{W_A}{l}}, l\in[1,2,4]$. and $N$ is the total number of patches of $i$th head, i.e., $N=T\times l^2, l\in[1,2,4]$.  $d_{head_{ij}}$ is the dimension of the $\mathbf{q_{i,m}}$. For each head $Head_i$,  we stack the partitioned patches  $\mathbf{q_{i}/k_{i}/v_{i}}$  from all frames of a video together to calculate the self-attention feature map of the video (see~\figureautorefname~\ref{fig:ViTransPAD}(b)), thus the self-attention of each head $Head_i$ always considers simultaneously the short attention focusing on local spatial information when $\mathbf{q_{i,m}/k_{i,n}}$ from the same frame of a video (see the red dotted line in~\figureautorefname~\ref{fig:ViTransAttent} denoting the short attention) and long attention capturing spatio-temporal dependencies over frames when $\mathbf{q_{i,m}/k_{i,n}}$ from the different frames (see the violet dotted line~\figureautorefname~\ref{fig:shortlongattent} denoting the long-range attention) . We can also jointly learn the spatio-temporal attention in a unified framework without learning the spatio-temporal attention independently as in~\cite{arnab2021vivit}.    
% \subsection{Over oss for face PAD}
% \label{sec:Loss}

\subsection{Loss function for face PAD}
Instead of adding a classification token to learn the representation of image as in ViT, we learn the representation of video based on all patch tokens without adding an extra classification token. In practice, the output of MLP Head servers as the classification embedding $\mathbf{Z}$ in this work (see~\figureautorefname~\ref{fig:ViTransPAD}). Then, the learned embedding $\mathbf{Z}$ is input in the cross-entropy loss function to train our model: 
\begin{equation}
\label{eq:cross_entropy} 
\mathcal{L}(\mathbf{Z};\Theta)= \sum^{K}_{k=1}-y_klogP(y_k=1|\mathbf{Z},\Theta) 
\end{equation}
where $\Theta$ are the parameters of model to be optimized and $K$ is the number of categories, i.e., $K=2$, which is the classes for face PAD being either bonafide or attack. 

\section{Experiments}
\label{sec:Experiments}
\subsection{Datasets and setup}
Datasets \textbf{OULU-NPU (O)}~\cite{Boulkenafet2017OULU}, \textbf{CASIA-MFSD (C)}~\cite{wen2015face}, \textbf{Idiap Replay Attack (I)}~\cite{ReplayAttack} and \textbf{MSU-MFSD (M)}~\cite{Zhang2012A} are used in our experiments. Attack Presentation Classification Error Rate (APCER),
Bona Fide Presentation Classification Error Rate (BPCER), Average Classification Error Rate (ACER) [20] and Half
Total Error Rate (HTER) [21] are used as evaluation metric in the intra/cross-datasets tests. In intra-dataset test, we follow the evaluation protocols of Oulu-NPU as in~\cite{Boulkenafet2017OULU}. In cross-datasets test, we conduct the evaluation on four datasets  \textbf{OULU-NPU (O)}, \textbf{CASIA-MFSD (C)}, \textbf{Idiap Replay Attack (I)} and \textbf{MSU-MFSD (M)}. We follow the OCIM protocols proposed in~\cite{shao2019multi} for cross-datasets test in which we randomly choose three of four datasets  as source datasets for training the model, and the remaining one is set as the target domain to evaluate the model. So, we have four experimental
modes: `O\&C\&I' to `M', `O\&M\&I' to `C', `O\&C\&M' to `I', `I\&C\&M' to `O'. 
%The proposed method is implemented with Pytorch with 3 NVIDIA Quadro RTX 6000 GPU. Models are trained with batch size 16 videos (8 frames per video) and Adam optimizer.

%OULU-NPU is a high-resolution database, containing four protocols to evaluate the generalization (e.g., unseen illumination and attack medium) of models respectively, which is used for intra testing. CASIA-MFSD, Replay-Attack, and MSU-MFSD are small-scale databases with low-resolution videos, which are used for domain generalization~\cite{shao2019multi} cross testing with a leaving-one-dataset-out protocol.

\subsection{Implementation Details}
%%\vspace{-0.2em}
\label{sec:Details}
All models are trained on 3 RTX 6000 GPUs with an initial learning rate of 1e-5 for 200 epochs
following the cosine schedule (50 epochs for warmup). Adam
optimizer and a mini-batch size of 16 videos (8 frames per
video sampled uniformly or with random interval) are applied
during training. Data augmentations including horizontal flip
and color jitter are used. 224x224 facial images cropped by
MTCNN [25] are used for both training and testing models.

\subsection{Ablation study}
All ablation studies are conducted on the Protocol-2 (different displays
and printers between training and testing sets) of OULU-NPU dataset unless otherwise specified.

%%%%%%%%%%%%%%% Ablation study CNN+Transformers %%%%%%%%%
%\vspace{0.3em}
\noindent\textbf{Effectiveness of convolutions in transformer}.\quad demonstrates a good computation-accuracy balance (ACER 1.19\% with GFLOPs 7.88) comparing to the pure CNNs or transformer (ViT/ViT(P)), which shows the effectiveness in introducing convolutions into transformers for face PAD. Comparing to convolutional token embedding, ViT is less effective for tokenizing patches (see ViT+T/ViT(P)+T).
%such as ViT either being pretrained (1.57\%) or not (7.25\%). However, the pure CNNs (here using EfficientNet to conduct convolutional patch embedding) without the self-attention are also inferior to our model. We also try to use pure transformer such as ViT to replace convolutions to employ patch embedding. The results (8.99\% or 3.38\%) shows again the effectiveness of convolutions for patch embedding.
The comparison experiments also show that the pretraining (denoted as P) is quite important for face PAD especially for the transformer-based architectures. 
\begin{table}[t]\small
\centering

 %\vspace{-1.0em}
\caption{Ablation study on effectiveness of convolutions introduced in transformer (P- Pretrained, T-Transformer).}
%%\vspace{-0.8em}
\scalebox{0.7}{
    \begin{tabular}{l|c | c|c|c|c}
        \toprule[1pt]
        Model & APCER(\%) & BPCER(\%) & ACER(\%) &\#GFLOPs&\#Params\\
        \hline
        CNN\tnote{a} & 5.43  & 1.08 & 3.26 &0.63&65M\\
        CNN (P) &2.12 & 3.30& 2.71 &0.63&65M\\
        \hline
        Vit & 10.30  & 4.20 & 7.25 &55.5&86M\\
        Vit (P) &2.38 & 0.76& 1.57 &55.5&86M\\
        \hline
        Vit+T & 11.32 & 6.67 & 8.99  &134.39&87M \\
        Vit (P)+T & 5.82 &0.94  & 3.38 &134.39&87M\\
        \hline
        CNN+T & 14.60 & 7.12 & 10.86   & 7.88 & 66M\\
        \textbf{CNN (P)+T} & 1.98 &0.40  & \textbf{1.19} &7.88 & 66M\\

        \bottomrule[1pt]
    \end{tabular}}
    % \begin{tablenotes}
    %     \footnotesize
    %     \item[a] CNN is EfficientNet in this work.
    % \end{tablenotes}

%\end{threeparttable}

\label{tab:CNNTrans}
\end{table}

%%%%%%%%%%%%%%%%%% MsMHSA %%%%%%%%%%%%%%%%%%%%
%\vspace{0.3em}
\noindent\textbf{Effectiveness of multi-scale MHSA}.\quad 
From Table~\ref{tab:MsMHSA}, we can see that all of the results using multi-scale patches  are better than the ones using single-scale patch. 
The best result is achieved when using multi-scale patches of size $28\times28$ and $14\times14$. However, the performance degrades when applying all three patches, which may be due to the overfitting on relative small dataset.  

%% Ablation study on Patch size and patch pyramids
\begin{table}[t]\small
\centering
 %\vspace{-1.0em}
\caption{Ablation study on the proposed Multi-scale MHSA.}
\scalebox{0.85}{
    \begin{tabular}{c|c|c|c}
        \toprule[1pt]
         28*28 & 14*14 & 7*7   & ACER(\%)\\
        \hline
         $\surd$  &        &         & 2.81 \\
                  &$\surd$ &         & 2.60\\
                  &        & $\surd$ & 2.32   \\
         $\surd$  & $\surd$&         & \textbf{1.19}\\
         $\surd$  &        &$\surd$  & 2.06\\
                  & $\surd$&$\surd$  & 2.12\\
         $\surd$& $\surd$&$\surd$  & 2.30\\
        \bottomrule[1pt]
    \end{tabular}}
%\vspace{-1.0em}
\label{tab:MsMHSA}
\end{table}

\begin{figure}[!h]
\centering
  \includegraphics[scale=0.55]{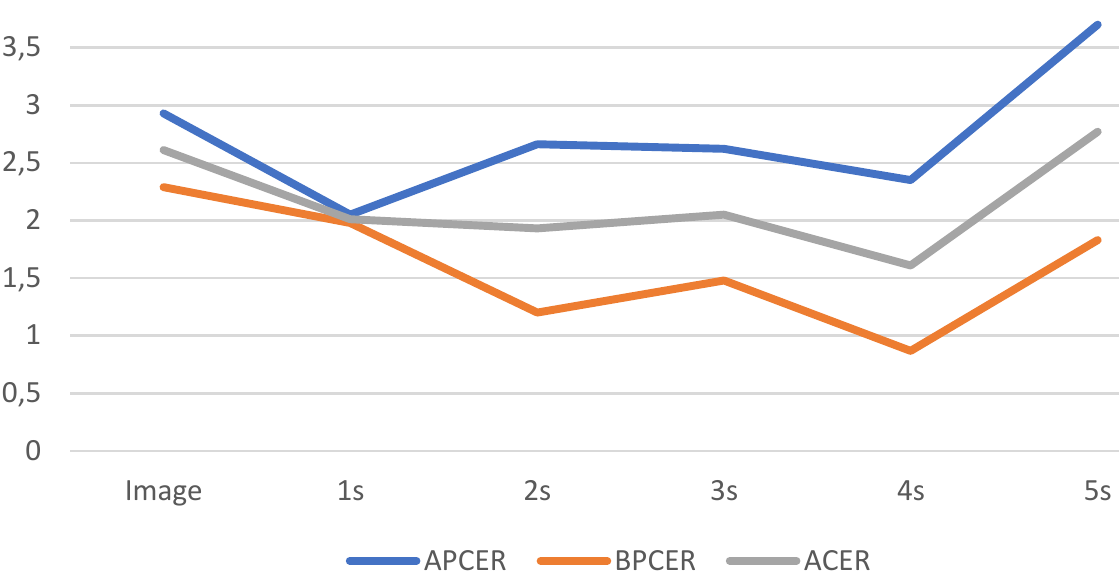}
  %\vspace{-1.0em}
  \caption{Study on short/long-range spatio-temporal attention.}
  \label{fig:shortlongattent}
\end{figure}
%%%%%%%%%%%%%% Image/Video attention %%%%%%%%%%%%%%%%%%%%
%\vspace{0.3em}
\noindent\textbf{Short/long-range spatio-temporal attention}. From~\figureautorefname~\ref{fig:shortlongattent}, we can see that the short spatial attention within an image within an image (the start of the grey curve) performs much worse than long spatio-temporal attention over frames spanning a video of 4 seconds in terms of ACER for face PAD. 
The more frames we use the better performance we gain until the length of video attains 4 seconds. 
This shows that the model 
%JCB 
%has been severely overfitting 
%
strongly over-fitted the
noise when the video is long (25fps$\times$4s).

\subsection{Intra-/cross-dataset Testing}
% As shown in Table~\ref{tab:OULU}, our proposed ViTransPAD first on the first three protocols (0.4\%, 1.3\% and 1.9\% ACER, respectively), which indicates the proposed method performs well at the generalization of the external environment, attack mediums and input camera variation. It is clear that the proposed ViTransPAD consistently outperforms CDCN~\cite{yu2020searching} on all protocols with -0.6\%, -0.2\%, -0.4\%, and -2.9\%, respectively, indicating the superiority of ViTransPAD. In the most challenging Protocol 4, ViTransPAD also achieves comparable performance with state-of-the-art SpoofTrace~\cite{liu2020disentangling} but more robustness with smaller ACER standard deviation among six kinds of unseen scenarios.
Table~\ref{tab:OULU} and Table~\ref{tab:DG} compare the performance of our method with the state-of-the-art methods on OULU-NPU~\cite{Boulkenafet2017OULU} and cross-dataset~\cite{shao2019multi} protocols. 
We can see that our proposed method ranks first on most protocols of OULU-NPU (i.e., Protocols 1, 2, and 4) and the `O\&C\&I to M', `I\&C\&M to O' of cross-dataset testing (`O\&C\&M to I' is the second best closing to the first).
%(i.e.,, [O\&C\&I to M] and [O\&C\&M to I]) without using domain knowledge for training,indicating the effectiveness and generalization of the learned spatio-temporal representation in unseen domain scenarios. 
Please note that our ViTransPAD can be also integrated in the framework of meta-learning as used in ANRL~\cite{liu2021adaptive} to further improve the cross-dataset generalization ability. The superior performance shows that our ViTrans can server as an effective new backbone for face PAD.
%, which is one of our future works. 

\begin{table}[t]
\centering

% %\vspace{-1.0em}
\caption{The results of the evaluation on the OULU-NPU dataset. Best results are marked in \textbf{bold} and second best in \underline{underline}.}
\resizebox{0.4\textwidth}{!}{
\begin{tabular}{c|c|c|c|c}
\toprule[1pt]

Prot. & Method & APCER(\%) & BPCER(\%) & ACER(\%) $\downarrow$ \\
\hline
\multirow{6}{*}{1}
        %&GRADIANT ~\cite{boulkenafet2017competition}&1.3 &12.5 & 6.9 \\
        %&MILHP &8.3 &0.8 & 4.6 \\
        %&TSCNN~\cite{chen2019attention} & 5.1 & 6.7 & 5.9 \\
        %&DRL-FAS~\cite{cai2020drl} & 5.4 & 4.0 & 4.7 \\
        %&CIFL~\cite{chen2021camera} & 3.8 & 2.9 & 3.4 \\
        %&STASN~\cite{yang2019face} &1.2 &2.5 & 1.9 \\
        &Auxiliary~\cite{Liu2018Learning} &1.6 &1.6 & 1.6 \\
        %&FaceDs~\cite{jourabloo2018face} &1.2 &1.7 & 1.5 \\
        &SpoofTrace~\cite{liu2020disentangling} &0.8 &1.3 & 1.1 \\
        %&Disentangled~\cite{zhang2020face} &1.7 &0.8 & 1.3 \\
        &FAS-SGTD~\cite{wang2020deep} &2.0 &0.0 & 1.0 \\
        &CDCN~\cite{yu2020searching} &0.4 &1.7 & 1.0 \\
        %&BCN~\cite{yu2020face} &0.0 &1.6 & 0.8 \\
        &DC-CDN~\cite{yu2021}& 0.5
       &0.3 &\underline{0.4} \\
   
        %&DeepPix~\cite{george2019deep}&0.8 &0.0 & \textbf{0.4} \\
        &\textbf{ViTransPAD (Ours)}&0.4 
       &0.2 &\textbf{0.3} \\

\midrule[1pt]
\multirow{6}{*}{2} 
       %&DeepPix~\cite{george2019deep}&11.4 &0.6 & 6.0 \\
       %&MILHP &5.6 &5.3 & 5.4 \\
       %&TSCNN~\cite{chen2019attention} & 7.6 & 2.2 & 4.9 \\
       %&FaceDs~\cite{jourabloo2018face}&4.2 &4.4 & 4.3 \\
       &Auxiliary~\cite{Liu2018Learning}&2.7 &2.7 & 2.7 \\
       %&Disentangled~\cite{zhang2020face} &1.1 &3.6 & 2.4 \\

       %&CIFL~\cite{chen2021camera} & 3.6 & 1.2 & 2.4 \\
       %&STASN~\cite{yang2019face}&4.2 &0.3 & 2.2 \\
       %&BCN~\cite{yu2020face} &2.6 &0.8 & 1.7 \\
       &SpoofTrace~\cite{liu2020disentangling} &2.3 &1.6 & 1.9 \\
       &FAS-SGTD~\cite{wang2020deep} &2.5 & 1.3 & 1.9 \\
       %&DRL-FAS~\cite{cai2020drl} & 3.7 & 0.1 & 1.9 \\
       &CDCN~\cite{yu2020searching} &1.5 &1.4 & 1.5 \\
       &DC-CDN~\cite{yu2021} &0.7 &1.9 &\underline{1.3} \\
  
       &\textbf{ViTransPAD (Ours)} &2.0 &0.4 &\textbf{1.2} \\
\midrule[1pt]
\multirow{4}{*}{3} 
       %&DeepPix~\cite{george2019deep}&11.7$\pm$19.6 &10.6$\pm$14.1 & 11.1$\pm$9.4 \\
        %&TSCNN~\cite{chen2019attention}&3.9$\pm$2.8 &7.3$\pm$1.1  &5.6$\pm$1.6 \\
       %&FaceDs~\cite{jourabloo2018face}&4.0$\pm$1.8 &3.8$\pm$1.2 &3.6$\pm$1.6 \\
      %&DRL-FAS~\cite{cai2020drl}&4.6$\pm$3.6 &1.3$\pm$1.8  &3.0$\pm$1.5 \\
       &Auxiliary~\cite{Liu2018Learning}&2.7$\pm$1.3 &3.1$\pm$1.7 &{2.9}$\pm$1.5 \\
       %&STASN~\cite{yang2019face}&4.7$\pm$3.9 &0.9$\pm$1.2  &2.8$\pm$1.6 \\
       &SpoofTrace~\cite{liu2020disentangling}&1.6 $\pm$1.6 & 4.0$\pm$5.4  &2.8$\pm$3.3 \\ &FAS-SGTD~\cite{wang2020deep}&3.2$\pm$2.0 & 2.2$\pm$1.4 & 2.7$\pm$0.6 \\
       %&BCN~\cite{yu2020face}&2.8$\pm$2.4 &2.3$\pm$2.8  &2.5$\pm$1.1 \\
      %&CIFL~\cite{chen2021camera}&3.8$\pm$1.3 &1.1$\pm$1.1  &2.5$\pm$0.8 \\
      &CDCN~\cite{yu2020searching} &2.4$\pm$1.3 &2.2$\pm$2.0  &2.3$\pm$1.4 \\
      %&Disentangled~\cite{zhang2020face}&2.8$\pm$2.2 &1.7$\pm$2.6  &2.2$\pm$2.2 \\
      &DC-CDN~\cite{yu2021} &2.2$\pm$2.8 &1.6$\pm$2.1  &\textbf{1.9$\pm$1.1} \\
      
      &\textbf{ViTransPAD (Ours)} &3.1$\pm$3.0 &1.0$\pm$1.3  &\underline{2.0$\pm$1.5} \\

\midrule[1pt]
\multirow{4}{*}{4} 
        %&DeepPix~\cite{george2019deep}&36.7$\pm$29.7 &13.3$\pm$14.1 & 25.0$\pm$12.7 \\
     %&TSCNN~\cite{chen2019attention}&11.3$\pm$3.9 &9.7$\pm$4.8  &9.8$\pm$4.2 \\
       &Auxiliary~\cite{Liu2018Learning}&9.3$\pm$5.6 &10.4$\pm$6.0 &9.5$\pm$6.0 \\
       %&STASN~\cite{yang2019face}&6.7$\pm$10.6 &8.3$\pm$8.4  &7.5$\pm$4.7 \\
     %&DRL-FAS~\cite{cai2020drl}&8.1$\pm$2.7 &6.9$\pm$5.8  &7.2$\pm$3.9 \\
       &CDCN~\cite{yu2020searching} &4.6$\pm$4.6 &9.2$\pm$8.0  &6.9$\pm$2.9 \\
     %&CIFL~\cite{chen2021camera}&5.9$\pm$3.3 &6.3$\pm$4.7  &6.1$\pm$4.1 \\
       %&FaceDs~\cite{jourabloo2018face}&1.2$\pm$6.3&6.1$\pm$5.1 &5.6$\pm$5.7 \\
       %&BCN~\cite{yu2020face}&2.9$\pm$4.0 &7.5$\pm$6.9  &5.2$\pm$3.7 \\
       &FAS-SGTD~\cite{wang2020deep}&6.7$\pm$7.5 &3.3$\pm$4.1 & 5.0$\pm$2.2 \\
       %&Disentangled~\cite{zhang2020face}&5.4$\pm$2.9 &3.3$\pm$6.0  &4.4$\pm$3.0 \\
     &DC-CDN~\cite{yu2021} &5.4$\pm$3.3 &2.5$\pm$4.2  &4.0$\pm$3.1 \\
     &SpoofTrace~\cite{liu2020disentangling}&2.3$\pm$3.6 &5.2$\pm$5.4  &\underline{3.8$\pm$4.2} \\

       &\textbf{\textbf{ViTransPAD (Ours)}} &4.4$\pm$4.8 &0.2$\pm$0.6  &\textbf{2.3$\pm$2.4} \\
\bottomrule[1pt]
\end{tabular}
}%\resizebox{\textwidth}{!}{
%%\vspace{-1.0em}
%%\vspace{1.0em}
\label{tab:OULU}
\end{table}

% %\subsection{Cross-dataset Testing}
% Table~\ref{tab:DG} compares the performance of our method with the state-of-the-art methods on DG cross-dataset protocols. We can see that our proposed method ranks first on most protocols (i.e.,, [O\&C\&I to M] and [O\&C\&M to I]) without using domain knowledge for training,indicating the effectiveness and generalization of the learned spatio-temporal representation in unseen domain scenarios. Please note that the performance of our ViTransPAD might be further improved when combining with DG-based learning strategies~\cite{shao2019multi,wang2021self,wang2020cross,jia2020single,shao2020regularized}, which is one of our future works. 

\begin{table}[t]
	\centering	
	\small

	 %\vspace{-1.0em}
	 		\caption{The results of cross-dataset testing protocol on OULU-NPU, CASIA-MFSD, Replay-Attack, and MSU-MFSD. }
	%The methods in the upper part are trained without domain information while those in the lower part are leveraging domain clues for generalization.
	
	%\caption{Results of cross-dataset testings on OULU-NPU, CASIA-MFSD, Replay-Attack, and MSU-MFSD with \textbf{supervised} setting. The methods in the upper part are trained without domain information while those in the lower part leveraging domain clues for generalization.}
	
	\scalebox{0.7}{\begin{tabular}{c|c|c|c|c}
		%\hline
	\midrule[1pt]
		\multirow{2}{*}{\textbf{Method}} & \textbf{O\&C\&I to M} & \textbf{O\&M\&I to C} 
		& \textbf{O\&C\&M to I}& \textbf{I\&C\&M to O}	
		\\
		
		& HTER(\%)  & HTER(\%)  & HTER(\%)  & HTER(\%) \\ \hline
		%MS\_LBP~\cite{Maatta2011Face}	              &29.76&78.50	&54.28&44.98	&50.30&51.64 	&50.29&49.31\\	
		%Binary CNN~\cite{yang2014learn}				  &29.25&82.87  &34.88&71.94 	&34.47&65.88 	&29.61&77.54\\
		%IDA~\cite{wen2015face}						  &66.67&27.86	&55.17&39.05	&28.35&78.25	&54.20&44.59\\
		Color Texture~\cite{Boulkenafet2017Face}      &28.09  &30.58    &40.40    &63.59 \\
		LBP-TOP~\cite{de2014face}    				  &36.90  &42.60    &49.45    &53.15 \\		
		Auxiliary(Depth)~\cite{Liu2018Learning}  &22.72	&33.52	&29.14	&30.17 \\
		%Auxiliary(All)~\cite{Liu2018Learning}		  &--&--		&28.4&--	    &27.6&--	    &--&--\\
		
		%\textbf{EPCR(Ours)}               				  &\textbf{12.5}&\textbf{95.3} 	&\textbf{18.9}&\textbf{89.7}	&\textbf{14.0}&\textbf{92.4}		&\textbf{17.9}&\textbf{90.9} \\

		%\hline
		
		MMD-AAE~\cite{li2018domain}					  &27.08 	&44.59	&31.58	&40.98 \\			
		MADDG~\cite{shao2019multi}   				  &17.69 	&24.5	    &22.19 	&27.98 \\
		MDRL~\cite{wang2020cross}                     &17.02 	&\underline{19.68}	&20.87	&25.02 \\
		SSDG-M~\cite{jia2020single}                   &{16.67} 	&23.11	&{18.21}	&25.17 \\
% 		SDA~\cite{wang2021self}                       &15.4&91.8 	&24.5&84.4	    &15.6&90.1	    &23.1&84.3 \\
% 		RFM~\cite{shao2020regularized}                &13.89&93.98 	&20.27&88.16	&17.3&90.48	    &16.45&91.16 \\
% 		D$^{2}$AM~\cite{chen2021generalizable}                       &12.70&95.66 	&20.98&85.58    &15.43&91.22	    &15.27&90.87 \\
		
% 		DRDG~\cite{liu2021dual}                       &12.43&95.81 	&19.05&88.79    &15.56&91.79	    &15.63&91.75 \\
		
% 		ANRL~\cite{liu2021adaptive}                       &10.83 & 96.75 	& 17.85 & 89.26	    & 16.03&91.04	    &15.67&91.90 \\		

        ANRL~\cite{liu2021adaptive}                       &\underline{10.83}  	&\textbf{17.85}  	    & \textbf{16.03}	    &\underline{15.67} \\

		\textbf{ViTransPAD (Ours)}               				  &\textbf{8.39} 	&21.27	&\underline{16.83}	&\textbf{15.63} \\
		%\hline		
		\bottomrule[1pt]
	\end{tabular}}

    \label{tab:DG}
    %%\vspace{-0.5em}
\end{table}

\subsection{Visualization}
\label{sec:visualisation}
\begin{figure}[!h]
\centering
\begin{minipage}{.2\textwidth}
  \centering
  %\subfloat[Liveness]{
  \includegraphics[width=\linewidth]{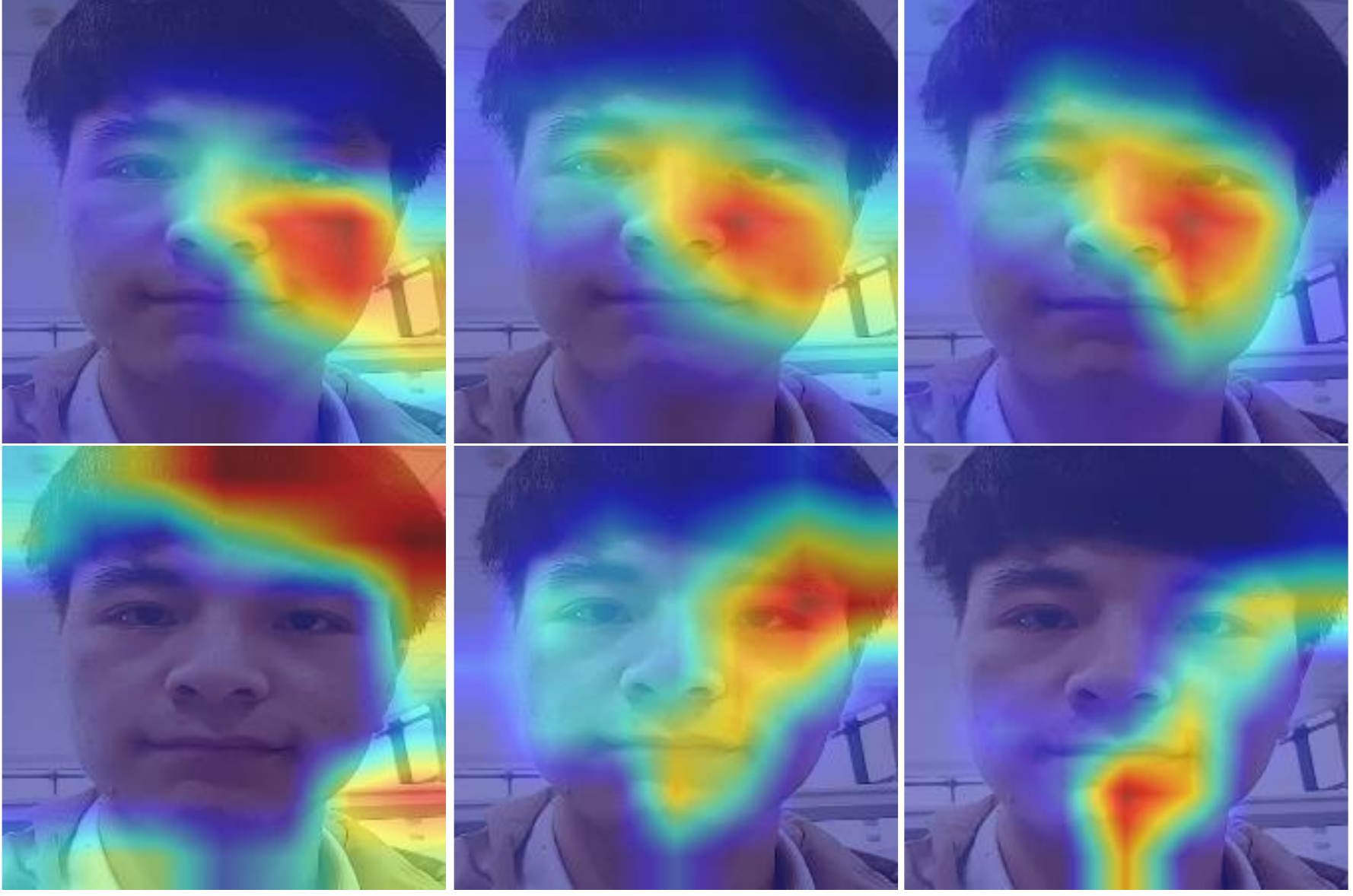}
  %\vspace{-1.6em}
  %}\quad
  %\subsection{}
  \label{fig:liveness_visualisation}
  \subcaption{Liveness}
  
\end{minipage}
\begin{minipage}{.2\textwidth}
  \centering
  %\subfloat[Attack]{
  \includegraphics[width=\linewidth]{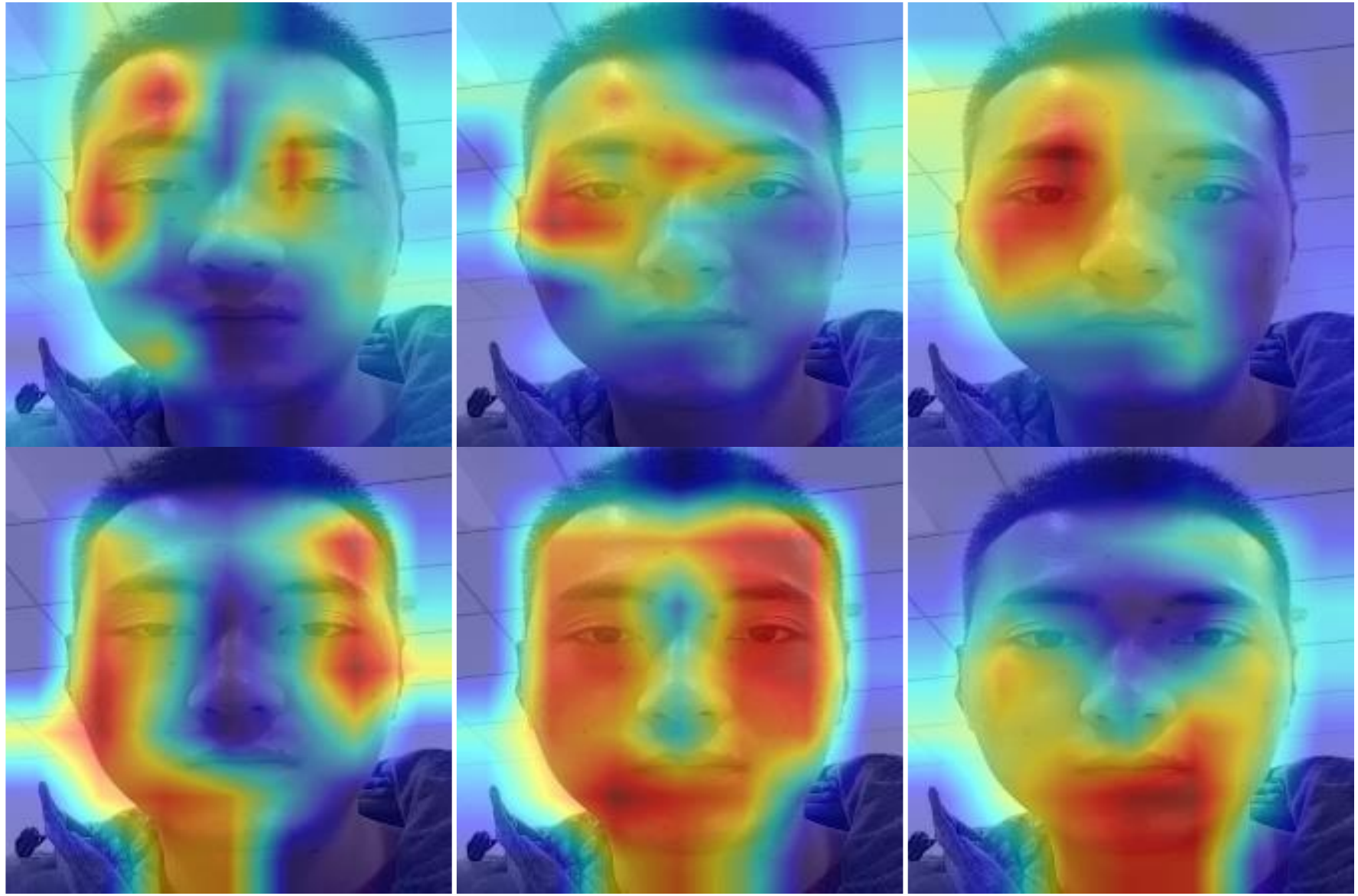}
  %\vspace{-1.6em}
  %}\quad
  \label{fig:attack_visualisation}
  \subcaption{Attack}
\end{minipage}
%\vspace{-0.8em}
\caption{Attention maps obtained by GradCAM~\cite{selvaraju2017grad}. Upper row illustrates attention maps of long-range attention over frames and the bottom row shows the ones of short-range attention within a frame.}
\label{fig:visualization}
\end{figure}

Due to lack the long-range spatio-temporal dependencies over frames, the short-range attention within a frame is vulnerable to the noise which results its attention maps are less consistent than the ones of long-range attention either for the liveness or attacks detection as shown in ~\figureautorefname~\ref{fig:visualization}. For instance, the upper row of (a) shows that the long-range attention always focus on the left half faces for liveness detection. However, the first attention map of short-range attention (the bottom row of (a)) attends the hairs on the forehead but the succeeding  frames switch to focus on the left half faces. The same results can be also observed in the attacks detection as shown in (b). 

% As shown in~\figureautorefname~\ref{fig:visualization}, the attention maps of long-range spatio-temporal attention over frames are more consistent than the ones of short-range spatial attention in a frame. In the detection of the liveness as shown in (a), the long-range attention always focus on the region of left half face (upper row of (a)). However, the attention region of short attention is not consistent even in the three adjacent frames (bottom row of (a)), in which the first frame attends the region of the hair but the succeeding two frames focus basically on the left half face. We can observe the same results in the detection of the video attack as shown in (b). The lack of the long-range spatio-temporal dependencies over frames results in the inconsistency of the attention maps obtained by the short attention within frame.     
% % The attention map is obtained by GradCAM~\cite{selvaraju2017grad}.
% \begin{figure}[!h]
% \centering
% \begin{minipage}{.2\textwidth}
%   \centering
%   \subfloat[Liveness]{
%   \includegraphics[width=\linewidth]{visualisation_cropped.pdf}}\quad
%   %\caption{Independent image branch}
% \end{minipage}
% \begin{minipage}{.2\textwidth}
%   \centering
%   \subfloat[Attack]{
%   \includegraphics[width=\linewidth]{visualisation_attack_cropped.pdf}}\quad
%   %\caption{Independent text branch}
% \end{minipage}
% %\vspace{-0.8em}
% \caption{Attention visualization. Upper row is the attention map of long-range attention over frames and the bottom row is the short-range attention in a frame.}
% \label{fig:visualization}
% \end{figure}

%\vspace{-1.2em}
\section{Conclusion}
%\vspace{-0.5em}
We design a Video-based Transformer for face PAD with short/long-range spatio-temporal attention which  can  not  only  focus  on  local  details  but also the context of a video. The proposed Multi-scale Multi-Head Self-Attention enables the model to learn a fine-grained representation to perform pixel-level discrimination required by face PAD. We also introduce convolutions to our ViTransPAD to integrate desirable proprieties of CNNs which can gain a good computation-accuracy balance. To the best of our knowledge, this is the first approach using video-based transformer for face PAD  which can serve as a  new backbone for further study.
% -------------------------------------------------------------------------
\bibliographystyle{IEEEbib}
\bibliography{strings}

\end{document}